\documentclass{article}

\usepackage[utf8]{inputenc}
\usepackage[backend=biber]{biblatex}
\usepackage{url}
\usepackage{graphicx}
\usepackage{tabularx}
\usepackage{multirow}
\usepackage{xspace}  
\usepackage[multiple]{footmisc}  
\usepackage{enumitem}
\usepackage{authblk}
\usepackage{subfig}
\usepackage{placeins}

\usepackage{xcolor}  

\title{\fbm:\\ Open, Transparent and Trusted Federated Learning for Real-world Healthcare Applications}
\author[1]{Francesco Cremonesi}
\author[2]{Marc Vesin}
\author[2]{Sergen Cansiz}
\author[2]{Yannick Bouillard}
\author[1]{Irene Balelli}
\author[1]{Lucia Innocenti}
\author[1]{Santiago Silva}
\author[1]{Samy-Safwan Ayed}
\author[1]{Riccardo Taiello}
\author[7]{Laetita Kameni}
\author[7]{Richard Vidal}
\author[4]{Fanny Orlhac}
\author[4]{Christophe Nioche}
\author[5]{Nathan Lapel}
\author[5]{Bastien Houis}
\author[5]{Romain Modzelewski}
\author[3]{Olivier Humbert}
\author[6]{Melek Önen}
\author[1]{Marco Lorenzi}
\affil[1]{Universit\'e C\^ote d'Azur, Inria Sophia Antipolis, Epione Research Group, France}
\affil[2]{Universit\'e C\^ote d'Azur, Inria Sophia Antipolis, SED Group, France}
\affil[3]{UCA, Universit\'e C\^ote d'Azur, France}
\affil[4]{Laboratoire d’Imagerie Translationnelle en Oncologie (LITO), U1288 Inserm, Université Paris-Saclay, Centre de Recherche de l’Institut Curie, Orsay, France}
\affil[5]{Nuclear Medicine Department, Henri Becquerel Center, Rouen, France}
\affil[6]{EURECOM, France}
\affil[7]{Accenture Labs, France}
\date{}

\newcommand{\fbm }{Fed-BioMed\xspace}

\begin{document}

\maketitle

\begin{abstract}
The  real-world  implementation  of  federated learning  is complex and requires research and development actions at the crossroad between different  domains  ranging from  data  science,  to software  programming,  networking, and security.  While today several FL libraries are proposed to data scientists and users, most of these frameworks are not designed to find seamless application in medical use-cases,  due to the specific challenges and requirements of working with medical data and hospital  infrastructures. Moreover, governance, design principles, and security assumptions of these frameworks are generally not clearly illustrated, thus preventing the adoption in sensitive applications.
Motivated by the current technological landscape of FL in healthcare, in this document we present \fbm: a research and development initiative aiming at translating federated learning (FL) into real-world medical research applications.  
We describe our design space, targeted users, domain constraints, and how these factors affect our current and future software architecture.
\end{abstract}

\section{Introduction}

The need for large amounts of data to develop  Artificial Intelligence (AI) in healthcare has motivated a number of national and international initiatives aimed at creating medical data lakes accessible to researchers, such as the French Health Data Hub~\cite{cuggia2019french}, the UK BioBank~\cite{sudlow2015uk}, the US ADNI~\cite{jack2008alzheimer} and TCGA~\cite{tomczak2015cancer}, among the many~\cite{sporns2005human,menze2014multimodal,clark2013cancer}. 
In spite of these initiatives, there are still major bottlenecks preventing the widespread availability of large centralized repositories of healthcare information~\cite{van2014systematic}. 

To overcome these limitations, Federated Learning (FL) has been proposed as a working paradigm to enable the training of ML models on large datasets from diverse sources while guaranteeing the respect of data privacy and governance. 
The basic paradigm of FL consists of iterating the following steps: i) model training is performed locally in the hospitals starting from a common initialization, ii) the resulting model parameters are subsequently shared (instead of the data) and aggregated, to define a global model iii) transmitted back to the hospitals to initiate a new local training step. 
Under certain conditions~\cite{mcmahan2016communication}, this procedure is guaranteed to converge to a final global model representing an optimal consensus among the hospitals participating in the experiment. 
FL is particularly suited for applications in sensitive domains, such as healthcare and biomedical research~\cite{rieke2020future,crowson2022systematic,darzidehkalani2022federated1}. 
The current societal and economical interest in FL for healthcare is paramount~\cite{shaheen2022applications, sadilek2021privacy}, as  demonstrated by the several large-scale medical research projects based on FL at the national and international level, focusing for example on rare hematological diseases\footnote{\url{https://genomed4all.eu/}}, drug development\footnote{\url{https://www.melloddy.eu/}}, blood cancer\footnote{\url{https://www.harmony-alliance.eu/}}, among the many~\cite{deist2020distributed,saldanha2022swarm,froelicher2021truly}. 

In spite of the current popularity, the real-world implementation of FL is complex and requires research and development actions at the crossroad between different domains spanning data science, software programming, networking, and security. 
Today, several FL software frameworks are currently being proposed to data scientists and users, based on different design spaces, goals, and with varying degrees of software maturity.
Nevertheless, most of these frameworks are not designed to find seamless application in medical use-cases, due to the specific challenges and requirements of working with medical data and hospital infrastructures. 
Furthermore, several widely-used FL libraries oftentimes devote little attention to describe the design space and guiding principles, while the spotlight is often placed on describing implementation details without justifying particular design choices.

In this document we present \fbm: a research and development initiative aiming at translating FL into real-world medical research applications, motivated by the current technological landscape of FL for healthcare.  
We describe our design space, targeted users, domain constraints, and how these factors affect our current and future software architecture.
While implementation details may change with the evolution of new technologies and contributions from a growing pool of developers, we believe that a description of \fbm's philosophy and guiding principles, along with the relevant architectural details, provides a faithful and transparent representation of our goal and ambitions.  

\subsection{Contribution}

While several FL frameworks are currently available, none of them have documented a clear set of design principles and guiding concepts inspired by the application domain, nor a demonstration of how their architecture and implementation satisfy the specific requirements of FL in clinical applications.
In what follows, we identify the requirements arising from the needs of medical data owners and biomedical data scientists, and show how \fbm addresses the specific challenges of this domain. 

To meet the strict security requirements typical of medical environments, \fbm focuses on empowering the user with a tight control of data management and model training process. 
This is based on a deployment workflow enabling easy setup of its main components, on prototypes of a data and model governance system with a graphical interface for non-technical users such as clinical data managers and physicians, on a Jupyter notebook interface for researchers and data scientists, and on extensive documentation and tutorials targeting biomedical data scientists as well as health data providers\footnote{\url{https://fedbiomed.gitlabpages.inria.fr/}}. 
Overall, the straightforward design of \fbm aims at simplifying the development and deployment of federated learning analysis in real-world healthcare research.

\section{Federated Learning for biomedical research applications: design considerations}\label{sec:FL-health}

The application of ML methods, and in particular of FL, in the context of medical data analysis presents unique challenges  requiring a targeted approach. 
First and foremost, is the conflict between the need for large datasets to train ML models and data sharing regulations, such as the European General Data Protection Regulation (GDPR)\footnote{\url{https://gdpr-info.eu/}} and the US Health Insurance Portability and Accountability Act (HIPAA)\footnote{\url{https://www.cdc.gov/phlp/publications/topic/hipaa.html}}. 
Additionally, ethical, economical, and technical barriers  hinder the sharing of medical information~\cite{van2014systematic,topol2019high}. 
The entities that provide medical records for FL experiments are therefore compelled to enforce strict data governance policies on their data, requiring strong security and privacy guarantees as well as retaining the ability to exert fine-grained control over data processing and flow.

Medical data are generally not directly amenable to FL analysis: they are often stored in unstructured, potentially proprietary formats inside independent silos, leading to large degrees of heterogeneity making data potentially biased and difficult to compare~\cite{hulsen2019big,topol2019high}.
For example, the lack of standardized coded definitions may lead to noisy labels, with a detrimental effect on training performance, especially for medical data where missingness is a common problem~\cite{ghassemi2020review}.
Big data in medicine is characterized by relatively low volumes with high information density~\cite{hulsen2019big}, and often requires integration of multiple data acquisition methods, bringing all the challenges associated with the analysis of multimodal data~\cite{kline2022multimodal}.
Furthermore, hospital IT infrastructure has been designed to support clinical and billing operations, but is not optimized for data analytics~\cite{hulsen2019big}. 
The procedures for installing and operating FL software may vary wildly across different hospitals, leading to difficulties during the deployment process and inconsistencies in the execution. 

In a biomedical research setting, the prototypical FL experiment consists of a dynamic and highly interactive series of training rounds interspersed with sessions of model and hyperparameter tuning, interpretation of partial results with domain experts, and general debugging. 
This process requires a large degree of interactivity that can be in contrast with the design of FL systems for other applications that are more focused on batch execution of training rounds, model stability, and high availability of the infrastructure.

Any software being operated in the context of biomedical research must satisfy strict security rules arising not only from data privacy concerns but also from intellectual property and compliance to guidelines~\cite{european2019ict}. 
The FL process itself exposes multiples vulnerabilities such as model inversion, membership inference, and model poisoning attacks~\cite{bouacida2021vulnerabilities,shyu2021systematic}, which must be mitigated through a combination of classical cybersecurity approaches --- such as e.g. encryption, firewalls, malware protection, and network segmentation--- as well as FL-specific techniques such as secure aggregation~\cite{mansouri2022learning}.

\subsection{Primary Requirements}\label{sec:requirements}

From the challenges described above we derive a set of requirements that \fbm aims at satisfying, categorized in four primary (i.e. must-have), four secondary (i.e. nice-to-have) and three minor (i.e. could-have) requirements. We summarize the challenges and requirements in Table~\ref{tab:requirements}.

\subsubsection*{Data and model governance} We define these as the granting the following rights to clinical data nodes: i) the ability to review, add or revoke at any time the availability of any given dataset for federated training; ii) the ability to approve, audit and monitor the execution of specific FL workflows; iii) the ability to review, audit and customize the deployment of the software infrastructure.  
Furthermore, as the time of data managers and clinical experts is often a scarce resource, the ability to exercise such fine-grained control on data governance and processing should be provided in a simple user interface requiring only minimal learning efforts.

\subsubsection*{Integration with biomedical data sources} An FL framework targeting interoperability should at the very least be aware of existing interoperability efforts in medical data management and analysis, and ideally should offer direct and seamless integration for the management of data in such formats. 

\subsubsection*{Researcher interactivity} FL frameworks suited for research purposes should provide the ability to launch, stop and generally manipulate the training process, modify model and training parameters on the fly, resume training from checkpoints, and monitor convergence, while respecting the FL paradigm, the data providers' privacy, and node's governance requirements. 

\subsubsection*{Security} An FL framework for the biomedical research domain should provide a secure environment for the FL technical infrastructure, minimizing the surface of attacks on the data providers' systems through the framework itself, implementing network segmentation, securing network communication, and insulating the execution from external attackers.
Furthermore, the framework should support and allow for easy activation of gradient protection, against model-targeted attacks such as model poisoning, membership inference, and model inversion.

\begin{table}
\vspace{0.5\baselineskip}
\begin{tabularx}{0.99\textwidth}{|
    >{\hsize=1.0\hsize\linewidth=\hsize\raggedright\arraybackslash}X
    | >{\hsize=1.4\hsize\linewidth=\hsize\raggedright\arraybackslash}X 
    | >{\hsize=0.6\hsize\linewidth=\hsize\raggedright\arraybackslash}X |}
    \hline
\textbf{Challenges}    & \textbf{Primary requirements}     & \textbf{Secondary and minor} \\[1ex] \hline\hline 
\begin{itemize}[itemsep=0.1pt, topsep=0pt, leftmargin=*]
    \item Data privacy and protection
    \item Siloed data
    \item Heterogeneous, unstructured data
    \item Integration of multiple data modalities
    \item Dynamic model development cycle
    \item Cyberattacks
    \item Model and gradient attacks
    \item Suboptimal IT infrastructure and computing capabilities
    \item IP protection
\end{itemize}
&
\vspace{1ex}
\textbf{Data and model governance}\newline
Powerful and easy-to-use governance tools for nodes.\vspace{0.3\baselineskip}\newline
\textbf{Integration with biomedical data sources}\newline
Support biomedical and interoperability standards; support diverse computing infrastructures.
\vspace{0.3\baselineskip}\newline
\textbf{Researcher interactivity}\newline
Support a highly interactive, dynamic model development cycle steered by the researcher.
\vspace{0.3\baselineskip}\newline
\textbf{Security}\newline
Secure software and related IT infrastructure against hacking; support model and gradient protection strategies.
\newline
&
\begin{itemize}[label={}, itemsep=0.1pt, topsep=0pt, leftmargin=*]
    \item federated pre/post processing
    \item support for several ML libraries
    \item portability
    \item drop-out tolerance
    \item high availability
    \item scalability
    \item lightweight
\end{itemize}
\\ \hline
\end{tabularx}
\caption{Summary of challenges and requirements addressed by fed-BioMed in the application of FL to biomedical research.}\label{tab:requirements}
\end{table}

\subsection{Secondary and Minor Requirements}

\textbf{Federated pre/post-processing}, i.e. the framework's ability to support data or model pre/post-processing in a federated approach.

\textbf{Supported ML libraries/frameworks}, meaning the framework's capability to seamlessly integrate with a variety of state-of-the-art ML libraries and providing state-of-the-art FL algorithms.

\textbf{Portability}, intended as reproducibility of the development environment as obtained e.g. by the use of containers and virtual environments.

\textbf{Drop-out tolerance}, as in the framework's resilience to node drop-outs, other unexpected failure events, or the framework's ability to provide fine-grained node selection during training. Note that in a controlled cross-silo setting unexpected drop-outs are usually considered less relevant than in a cross-device FL deployment.	


Finally, we note that some other requirements are often cited in the broader context of FL applications but are of relatively minor importance to the domain of biomedical research, such as e.g. high availability of servers, scalability in terms of number of clients, and lightweight software implementation.

\section{Background: FL Frameworks Landscape}\label{sec:frameworks}
The number of FL frameworks has dramatically surged in recent years, witnessing the growing interest in the applications of this technology~\cite{li2021survey,shaheen2022applications,ogundokun2022review}.
We are aware of well-established products whose focus is not related to biomedical applications, but whose implementation does not exclude future deployments in this domain, such as e.g. Tensorflow Federated (TFF)~\cite{bonawitz2019towards}, FedML~\cite{he2020fedml}, IBM-FL~\cite{ludwig2020ibm}, FATE~\cite{liu2021fate}, PaddleFL~\cite{ma2019paddlepaddle}, and PySyft~\cite{ryffel2018generic}.
Henceforth, we restrict our analysis to SubstraFL~\cite{owkin2022substrafl}, OpenFL~\cite{foley2022openfl}, Flare~\cite{roth2022flare}, and Flower~\cite{beutel2020flower}, which are frameworks that have already been applied in medical use-cases, in accordance with the focus of this paper. 
Table~\ref{tab:frameworks} summarizes our frameworks review in light of the requirements identified in section~\ref{sec:requirements}.

\begin{table}
\begin{small}
\begin{tabularx}{0.99\textwidth}{|
    >{\raggedright\arraybackslash}X
    | >{\raggedright\arraybackslash}X 
    >{\raggedright\arraybackslash}X 
    >{\raggedright\arraybackslash}X 
    >{\raggedright\arraybackslash}X|}
    \hline
     & \textbf{SubstraFL}    & \textbf{OpenFL}    & \textbf{Flare}   & \textbf{Flower} \\[1ex] \hline\hline 
Design focus                      & collaboration, privacy, \mbox{traceability}        & cybersecurity                                    & scalability, flexibility, lightweight          &  scalability, flexibility, agnostic to clients, communication and privacy      \\[8ex] \hline 
Data \mbox{governance}                   & Distributed ledger, RBAC, GUI               & HTTP API, Python SDK                             & configuration files, Python SDK                &   Python SDK                                   \\[8ex] \hline 
Model \mbox{governance}                  & Distributed ledger, RBAC, GUI               & None built-in                                    & None built-in                                  &   None built-in                                   \\[8ex] \hline 
Integration with medical domain   & Only \mbox{demonstrators}                          & Only \mbox{demonstrators}                        & Only \mbox{demonstrators}                      &   None built-in                                   \\[8ex] \hline 
Researcher \mbox{interactivity}          & Explicit focus on production                & Limited                                          & Limited                                        &    High interactivity through Python SDK                                  \\[8ex] \hline 
Cybersecurity                     & encrypted \mbox{communication}                     & TEE, \mbox{encrypted} \mbox{communication}, PKI                & encrypted \mbox{communication}                        &  encrypted communication                                    \\[8ex] \hline 
Model and \mbox{update} attacks          & Future development                          & None built-in                                    & Only demonstrators                     & Secure aggregation \\
\hline
\end{tabularx}
\end{small}
\caption{Comparison of FL frameworks with respect to the requirements identified in Section \ref{sec:requirements}. 
The label ``Only demonstrators'' signifies that a feature is not built in the framework, but that a demonstrator has been provided in the form of a reference implementation, a tutorial, or documentation.}
\label{tab:frameworks}
\end{table}


LabeliaLab's SubstraFL framework is a Python library based on the Substra software developed by the company Owkin~\cite{owkin2022substrafl,galtier2019substra}. 
It is currently being used in healthcare applications for drug discovery in the context of the Mellody project~\cite{owkin2019mellody}, as well as oncology, anatomopathology and fertility in the context of the Healthchain project~\cite{owkin2019healthchain, ogier2023federated}. 
SubstraFL's architecture is based on a fully-decentralized distributed ledger, and is designed upon the three core principles of collaboration, privacy, and traceability.  
Data governance is implemented through operations on the distributed ledger, which also guarantees traceability of all ML operations within the consortium.
Operational roles are well-defined, and a permissions regime inspired by the Role-Based Access Control paradigm (RBAC) can be used to enable additional governance measures for handling remote \emph{assets} (algorithms and data).
Furthermore, a Graphical User Interface (GUI) is provided to simplify the management of assets on nodes. 
While SubstraFL has been deployed in healthcare settings, to our knowledge no tools specifically dedicated to the management of healthcare or biomedical assets are provided with the library, nor are there any tutorials or deployment examples focused on this domain.
SubstraFL's target use-case is the execution of FL experiments at scale, and therefore the library is intended to be mainly used in production environments. 
For this reason, it may not be an ideal choice for exploratory workflows requiring a high degree of interactivity during model development and training.
The distributed ledger at the core of SubstraFL's architecture also acts as a security feature by providing a strong guarantee against the alteration of training and inference metadata, and potentially against model poisoning attacks. 
No explicit description of encrypted secured communications, such as TLS, is currently provided in the framework, while differential  privacy and secure aggregation approaches are not integrated in the library. Finally, governance and roadmap of SubstraFL are centralised, preventing the development of an open-source community around the project.

OpenFL is another Python-based FL library, originally designed for a healthcare application but later expanded to be use case-agnostic~\cite{foley2022openfl}.
It has been used in the context of medical applications for a global FL deployment aimed at detecting glioblastoma sub-compartment boundaries~\cite{pati2022federated}. 
OpenFL has been designed to support multi-institutional collaborations with a strong focus on cybersecurity. 
The system architecture is based on the star topology paradigm with the aggregator as the central node and collaborators as edge nodes, authenticated through PKI certificates and communicating via encrypted TLS connections. 
Governance is implemented mainly through API operations or code snippets written by data owners. 
For example, the code implementing a \emph{ShardDescriptor} needs to be present on each node.
In some setups, data scientists may query centralized datasets and experiment registries holding metadata describing the whole federation.
We could not find any core functionalities in the library that are specific to the integration with biomedical data sources. 
However, tutorials with a medical focus are provided in the examples section and can serve as a rough template for simple biomedical applications. 
From the point of view of interactivity, OpenFL's documentation and experiment API seems to implicitly emphasize batch execution of training experiments rather than model exploration, even though some features, such as e.g. the simple porting of model descriptions from the serial to the federated approach, do enable some degree of interactivity.
OpenFL has a strong declared focus on cybersecurity. 
The usage of hardware-level features such as Trusted Execution Environments (TEE), as well as more conventional network-level measures such as TLS-encrypted communication and PKI certificates guarantee high degree of protection against attacks. 
Nevertheless, fully exploiting the capabilities of OpenFL is bound to the adoption of proprietary hardware. 
This condition is not necessarily compatible with practical deployment of FL in hospitals, and may critically prevent the adoption of the software in typical real-world scenarios.

Flare is yet another Python FL library, developed by Nvidia~\cite{roth2022flare}, which has been recently used in a world-wide federation of clinical sites to develop a new model for triaging patients affected by COVID-19~\cite{dayan2021federated}, as well as other healthcare-related demonstrations for e.g. classification and segmentation tasks on medical images~\cite{roth2020federated,sarma2021federated}.
Flare has been designed on the principles of scalability, flexibility and lightweight, and is targeted towards cross-silo FL, not limited to healthcare, supporting both production settings as well as simulated FL for researchers.
Flare's architecture is built on the paradigm of one \emph{Controller} distributing \emph{tasks} to several \emph{workers}, thus leading to a more generic framework with respect to the other libraries analysed here.
Governance is handled through configuration files and a Python code managed manually by node administrators. 
Specifications for a specific training experiment are provided partially by the data scientist (e.g. the model) and partially by the clients (e.g. the learners).
To our knowledge, no special tools are provided to node administrators for the management of data nor for fine-grained control over algorithm execution.
Moreover, while Flare's utility has been showcased in healthcare settings, it still aims to be a generic framework, and as such we could not identify features of the library aimed specifically at the integration with biomedical data sources.
Similarly as with OpenFL, the capabilities of Flare are oriented towards the adoption of specialized hardware, as demonstrated by the focus of the project on GPU usage, scalability, and high availability, which in our experience does not correspond to a prototypical hospital deployment scenario.
Flare's reliance on configuration files leads to a somewhat complex ecosystem that could potentially lead to a more rigid structure, lacking some of the flexibility that medical researchers may wish for during an exploratory model development phase. 
From a cybersecurity point of view, Flare offers protection against man in the middle and impersonation attacks, through the use of encrypted communication, SSL authentication, and a robust provisioning workflow.
Furthermore, the concept of local \emph{filters} enables the implementation of secure aggregation, homomorphic encryption, and differential privacy, which are however not available as part of the standard features of the framework.

Recently, the  widely-used Flower framework~\cite{beutel2020flower} has announced a collaboration with the Swedish decentralized AI project in order to improve the national healthcare ecosystem\footnote{\url{https://flower.dev/conf/flower-summit-2022/}}.
The design principles at the basis of Flower's architecture are scalability, flexibility, and generality w.r.t ML, communication and privacy frameworks. 
Given the general focus, Flower does not come pre-equipped with any data governance tools, nor does it provide any specific tools for the integration with biomedical data formats. 
On the other hand, Flower's flexible structure makes it highly amenable to highly interactive workflows, which are also further customizable thanks to the extensible \texttt{Strategy} class.
Flower offers native support for encrypted communication through its use of gRPC~\cite{grpc2018high}.
Their Secure Aggregation implementation, named \texttt{Salvia}, remains to our knowledge a proof of concept that is not yet built into the library.

\section{Methods}

\subsection{Design space and goals of \fbm}\label{sec:design-space}

The main purpose of \fbm is to enable seamless collaboration between medical investigators, data providers, and data scientists in a high trust, highly interactive research environment. 
First and foremost, we aim at providing secure tools for the governance of personal biomedical data and the federated training of models on such data. 
Additionally, we strive to make all our interfaces, especially those facing the medical researchers and data providers, easy to use even for non-technical users. 
Our second goal is to make the interface for the data scientist as interactive and flexible as possible to allow fast turnaround during the development of new ML models and strategies, while respecting the privacy needs of the data providers. 

\begin{figure}
\centering
\includegraphics[width=0.99\textwidth]{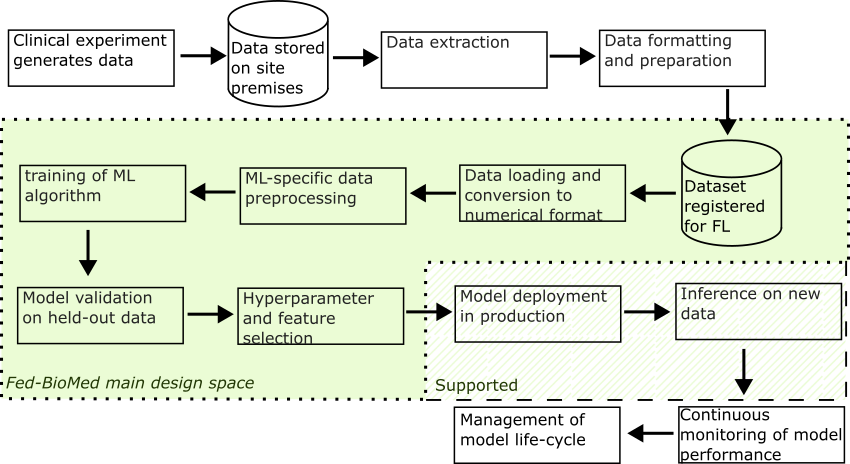}
\caption{Workflow of an FL experiment from the point of view of a clinical data provider. The shaded area represents \fbm's design space in terms of functionalities and targeted usage.
}
\label{fig:fl-workflow}
\end{figure}

In the context of the life cycle of a FL experiment, we make a clear-cut distinction between research applications and model deployment in production. 
Both begin with a data generation and preparation step, which is typically within the scope of a specific clinical investigation. 
In the case of secondary use of data, this step includes extracting data from an already existing hospital database, cleaning and wrangling the data, and applying the necessary anonymization or pseudonymization procedures.
This step is followed by a model development phase, comprising federated data preprocessing, training, validation, and hyperparameter tuning. 
Sometimes the model development phase is split in two sub-phases: an initial exploration where a subset of the data is centralized, followed by the actual federated training phase.
In the development phase, emphasis is put on the scientific process of hypothesis formulation, experiment and model design, architecture and hyperparameter search, and training.
Finally, once a stable version has been reached, the model can be deployed in production mode, where emphasis is placed more on inference, reliability, robustness and high availability.
The main scope of \fbm is to support the deployment and translation of AI to biomedical research and healthcare during the model development phase, but we also support deployments in production mode. 
To this end, we are actively working and collaborating with experts in medical data analysis, optimization, security, databases, and visualization. 
Figure~\ref{fig:fl-workflow} shows the typical workflow of a FL experiment, with the main design space of \fbm highlighted in the shaded area and the supported deployment modes highlighted in the dashed area. 

Our current design goals include using \fbm in a high-trust environment, where all parties are honest and have an open, secure channel of communication outside of it. 
We target mainly research consortia composed of a relatively small number of data providers (e.g. university hospitals or medical research centers) and data consumers (e.g. data scientists and researchers with an expertise on biomedical data). 
In the future, we aim to extend our paradigm to account for less-trusted environments, such as e.g. in the presence of malicious clients.

Federated learning requires the availability of computing infrastructure at the edge nodes: rather than focusing on high performance computing, we target the more common scenario of clinical sites with varying computing capabilities. 
We do not make the assumption that the edge nodes' computing infrastructure is able to sustain a high availability cycle; instead we consider that the research work will be organized in periods of active testing and model training, alternating with periods of independent work where the computing infrastructure at the edge nodes may be disconnected. 

During periods of research activity and model training, we assume that all actors involved will be available and able to communicate with low response latency. 
Our researcher and medical data provider interfaces are designed with this underlying assumption, thus enabling us to provide what we deem to be a good balance between automatizing some governance processes while keeping a human-in-the-loop approach to maximize security. 

The data scientists using \fbm, called \emph{researchers} in our notation, are assumed to be knowledgeable users of at least one of the supported ML frameworks, and to be able to read API documentation and code basic Python functions complying with it. 
We also assume that they are able to discuss with biomedical researchers and data providers about technical aspects involving both the preparation of data (e.g. developing a common data model and format) and the development or interpretation of ML models for biomedical data. 
Coherently with the FL paradigm, we make the assumption that the edge nodes wish to protect their data from arbitrary access of the researchers while preventing data leakages, from either a malicious source or a manipulation mistake. 

\subsection{Design and functional architecture}\label{sec:design-and-functional}

\begin{figure}
\centering
\includegraphics[width=0.99\textwidth]{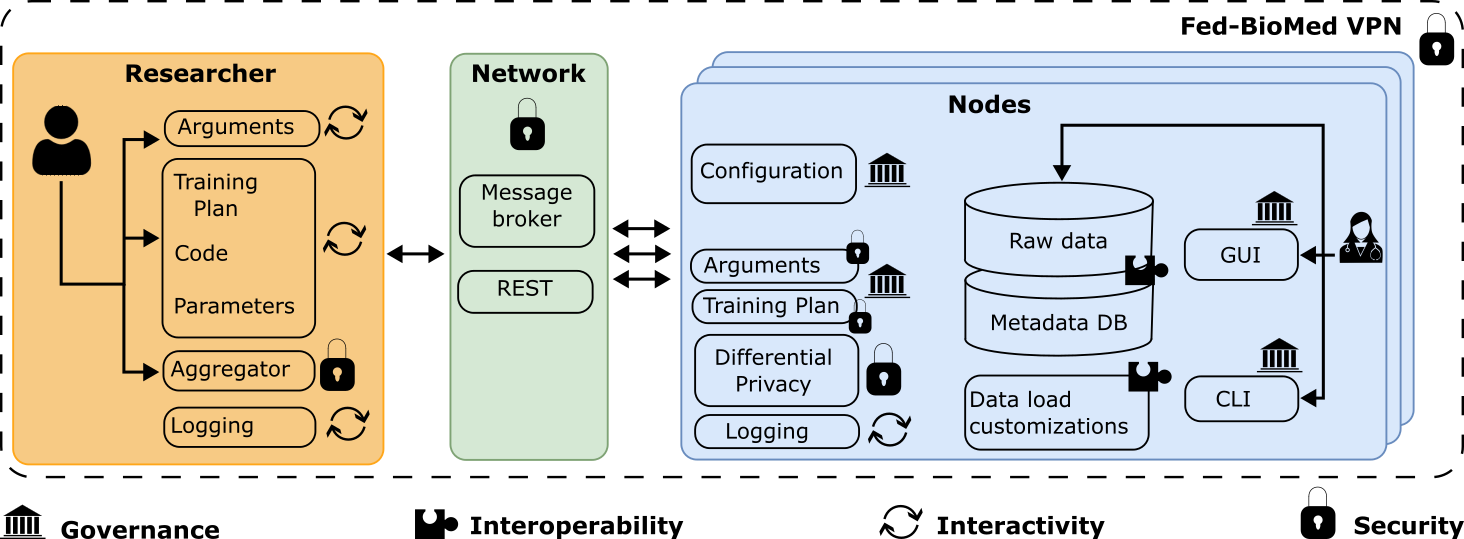}
\caption{High-level architecture and design pillars of \fbm: node-side governance and control, interoperability with medical data standards and infrastructure, researcher interactivity, and data privacy and security. \fbm is composed of three components: the researcher, a data scientist responsible for designing and steering the training of the ML model; the nodes, i.e. the clinical data providers; and the network, responsible for brokering all communication between the researcher and nodes. Each component has been designed following the Fed-BioMed requirements (Section 2), and the figure highlights which requirement affects the architectural subcomponents.}
\label{fig:fbm-pillars}
\end{figure}

The functional architecture of \fbm, as shown in Figure~\ref{fig:fbm-pillars}, is based on the design space described in Section~\ref{sec:design-space} and the requirements identified in Section~\ref{sec:FL-health}. 
The \fbm ecosystem comprises three architectural components: the network, the researcher, and one or more nodes.
The network is responsible for brokering the communication between all \fbm components, the researcher is responsible for configuring and driving the federated learning experiment, as well as aggregating the trained models, while the nodes are responsible for local data governance and actually performing the training. 

\begin{figure}
    \centering
    \includegraphics[width=\textwidth]{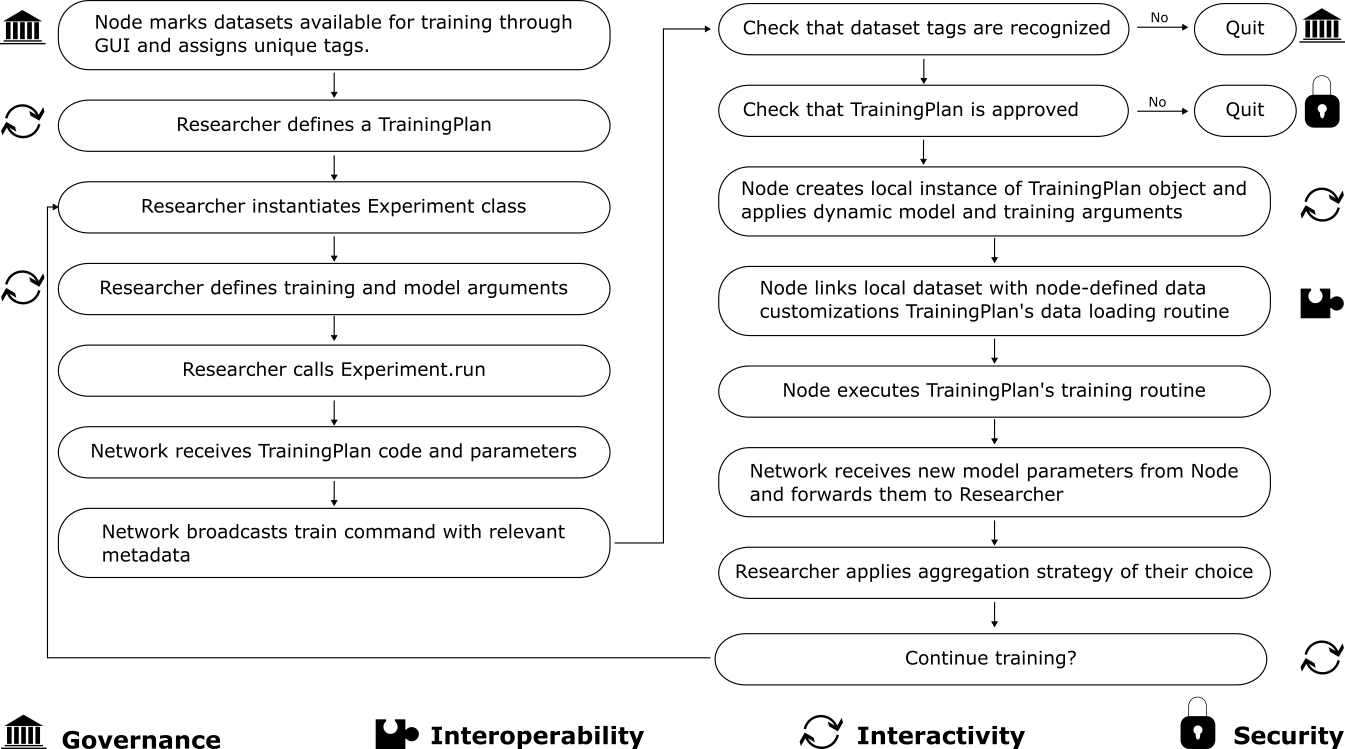}
    \caption{Prototypical FL training workflow in \fbm. Icons indicate the steps where the design was influenced by a particular requirement. First the nodes mark their dataset as available for federated training, while the researcher defines and obtains approval for a TrainingPlan. Then the researcher may launch multiple experiments, and within each experiment interactively launch multiple rounds of training. In each round, data are loaded locally on the nodes and the researcher-defined training routine is executed. Communication of model parameters and metadata always happens through the Network component.}
    \label{fig:exp-loop}
\end{figure}

A standard training experiment is described in Figure~\ref{fig:exp-loop}, where nodes are first required to make their data available for training by inserting an appropriate metadata entry in a locally-stored database, and assigning unique identifying tags to those data.
This process is simplified by the web-GUI built-in to our framework, but a CLI is also available for programmatic approaches.
Optionally, node-specific customizations to the data loading process may be specified here through a plugin system called \texttt{DataLoadingPlan}, with the intention of reducing the data formatting burden by providing a logical layer between the researcher and the actual data format as stored locally.
Then researchers define a \texttt{TrainingPlan}, i.e. an object containing the description of the ML model and aggregation parameters, the code for the training and validation routines, a data loading and preprocessing routine, and other training-specific information such as e.g. the description of the optimizer.
The \texttt{TrainingPlan} is packaged in an \texttt{Experiment} object along with the aggregator, training arguments, and data identifier tags. 

The researcher then issues a train command through the network's message broker, which is broadcasted to all the nodes.
Nodes that identify the requested data tags within their metadata database may begin training: first the \texttt{TrainingPlan} object is recreated on the node, then data are loaded, preprocessed using both node-specific customizations as well as researcher-specified functions, and finally the training routine of the training plan is executed.
For convenience, \fbm comes with pre-packaged \texttt{TrainingPlans} for widely-used frameworks, thus requiring minimal configuration from researchers.

\subsubsection*{Node-side governance}

\fbm empowers nodes by maximising the amount of control they have over the execution of a FL experiment.
In \fbm, there is no notion of a trusted third party to which nodes must delegate full trust.
For training, we make the assumption that datasets have already been treated to comply to existing data privacy regulations (e.g. pseudonymisation), and that they have been at least partially harmonised, both semantically and syntactically, to a pre-defined format common to all the nodes participating in the experiment.

When an experiment is launched, the datasets corresponding to the requested tags are loaded in the node's worker instance and prepared for training.
A researcher-defined \texttt{training\_data} function is then used to load the data and apply local pre-processing.  
Although this is not yet strictly enforced, this function is expected to use one of the dataset classes defined by \fbm providing controlled access to filesystem resources and best compatibility with the rest of the \fbm workflow. 
It is also possible to emulate a federated data pre-processing pipeline by assembling multiple experiments, thanks to \fbm's interactive approach.

\fbm's answer to the obvious tradeoff between allowing researchers to execute custom code snippets on the worker nodes and the associated security issues, is to provide a mechanism called \emph{training plan approval}. 
When enabled, this feature prevents the execution of code contained in \texttt{TrainingPlans} that has not yet been inspected and approved by the node.
A hash of the code is computed and checked at every training execution to prevent substitution attacks, and while this feature is not a definitive fail-safe measure against malicious users, it provides an additional layer of protection, as it empowers nodes and fosters researcher-node collaboration through shared responsibility.
Furthermore, nodes are granted the right to override certain training parameters, regardless of the researcher's original request, for matters concerning security and resource usage.

\subsubsection*{Integration with biomedical data sources}

Like most FL frameworks, \fbm does not offer a direct connection with raw data sources such as hospital EHR, PACS, and other clinical IT systems. 
Instead, clinical data managers are requested to perform a one-time data preparation task to extract and wrangle the data.
Contrary to other frameworks, \fbm tries to minimize this effort by implementing dedicated mechanisms and offering built-in integration with widely-used data standards, at the cost of losing some generality by restricting its focus on biomedical data. 

\fbm introduces the notion of a \texttt{DataLoadingPlan}, meaning a set of customizations configured by the node which allow changing the way data is presented to the researcher. 
Built-in \texttt{DataLoadingPlans} are already integrated in \fbm's GUI, thus providing a mechanism for data harmonization that does not require extensive effort on the node side.

\fbm also provides a built-in suite of dataset classes that provide an interface to common standards. 
For example, our \texttt{MedicalFolderDataset} class is inspired by the BIDS standard\footnote{\url{https://bids.neuroimaging.io/}} as well as PyTorch's \texttt{ImageFolder}.
Furthermore, our generic \texttt{TabularDataset} class supports any standard that may be reduced to csv format.
As part of our continuous development effort, we plan to significantly expand this suite to include several other healthcare interoperability standards, depending on the needs of our collaborators. 

Hospitals participating in the same FL experiment may have different computing infrastructures. 
\fbm supports execution on containers, virtual machines, bare-metal CPUs and GPUs in the same experiment, allowing for the reconciliation of an heterogeneous computing infrastructure across nodes.

\subsubsection*{Researcher interactivity}

The researcher-facing side of \fbm is a Python SDK for configuring and managing FL experiments, preferably via jupyter notebooks. 
To create an experiment, researchers first instantiate a class called \texttt{TrainingPlan} containing the definition of the federated data and the training routine. 
The code written by data scientists in the \texttt{TrainingPlan} is designed to be as similar as possible to the serial local version of the training loop. 
This interface is intentionally generic to support a wide variety of use cases, from workflows allowing to compute federated summary statistics to general distributed optimization based for example on Stochastic Gradient Descent (SGD), Expectation Maximization (EM), and Variational Inference (VI).
For convenience, we offer specialized \texttt{TrainingPlans} for specific optimization strategies such as SGD, and for the ML framework being used.  

Our approach promotes researcher interactivity via the \texttt{Experiment} class, which allows to easily set the participating nodes, the FL strategy, and includes the \texttt{TrainingPlan} itself.
The \texttt{Experiment} is provided with a logging functionality and integration with tensorboard~\cite{abadi2015tensorflow}.
The modular design of the \fbm training loop allows researchers to dynamically adjust hyper-parameters on the fly.
Furthermore, a checkpointing system allows saving and loading the state of an experiment in persistent memory. 

In a highly secure environment where training plans must be approved by each node, minor changes in a training plan may lead to insufferable delays for obtaining multiple approvals.
Therefore, we make a distinction between the \texttt{TrainingPlan} source code and a set of training and model arguments. 
The former must be approved by the node, but it may leverage the latter as a way to dynamically configure details about the experiment that lie within the node's acceptable ranges. 
For example, a \texttt{TrainingPlan} may include a dropout layer but the dropout rate would be specified as a model argument, thus providing security guarantees to the node while allowing flexibility for the researcher. 

\subsubsection*{Cybersecurity}

\fbm’s current threat model assumes that  nodes and server are honest-but-curious, nodes are independent actors and do not collude, and researchers may be malicious.
This set of assumptions corresponds to our current application scenarios in hospital networks, where clients are trusted and strong protection is required with respect to researcher’s manipulations.
Thus, the first measure that we put in place is deploying the whole \fbm  ecosystem inside a VPN, to effectively isolate the execution and protect it against external attacks.
In the honest-but-curious model, an entity that gains access to the model weights may still attempt to re-identify individual samples through model inversion attacks.
\fbm implements both differential privacy as well as secure aggregation based on additively homomorphic encryption~\cite{joye2013scalable,mansouri2022learning}, with currently some limitations concerning the supported ML frameworks and aggregation methods.

\fbm's architecture offers one final layer of protection, by insulating the researcher from the nodes through the existence of a middle component, the network, that brokers all communication between them. 
Furthermore, our \texttt{TrainingPlan} approval mechanism can also be viewed as a security feature, by allowing nodes to review any code intended to be executed within their systems. 

In  our  short-term  roadmap  we  plan  to  strengthen  our  security  features by  adopting  a  tighter  threat  model.
The  first  measure  that  we  intend  to introduce  is  encrypted  communication  inside  the  VPN, preventing  potentially malicious actors who have gained internal access from being able to listen-in on the communication exchanges.  
Secondly, we intend to introduce trusted digital certificates to improve our protection against impersonation attacks.


\section{Status report}
\subsection{Current state of the implementation and future plans}

\fbm is a constantly evolving project, with an active group of core maintainers striving to issue monthly releases.
The current maturity of our library has been acknowledged to allow a first real-world deployment and validation within a federation of some members from the UniCancer consortium, as described in Section~\ref{sec:applications}, thus characterizing its Technology Readiness Level as TLR 5\footnote{\url{https://ec.europa.eu/research/participants/data/ref/h2020/wp/2014_2015/annexes/h2020-wp1415-annex-g-trl_en.pdf}}.
However, we strive to constantly improve several aspects of the implementation in future releases, following a development roadmap inspired by the design pillars described in this paper and actively driven by the needs of our growing community of users. 
Building on our functional description in Section~\ref{sec:design-and-functional}, we provide in the Supplementary Information~\ref{si:technical} a fine-grained description of the core library, interface and cybersecurity functionalities currently implemented in \fbm.

\subsection{Demonstration on real-world federated hospitals networks}\label{sec:applications}


To illustrate the capabilities of Fed-BioMed for real world applications of federated learning in hospitals networks, in this section we report the results obtained from the end-to-end deployment and training of a federated prostate segmentation model, based on the data hosted by three different medical institutes from the Unicancer consortium \footnote{\url{https://www.unicancer.fr/en/}}. 

The French hospitals involved are the \emph{Centre Henri Becquerel}, Rouen, the \emph{Institut Curie}, Orsay, and the \emph{Centre Antoine Lacassagne}, Nice. Each hospital loaded into the Fed-BioMed client the respective dataset composed by prostate magnetic resonance images (MRIs), and associated prostate segmentation masks. The dataset considered for this test were the following:

\begin{itemize}
   \item \textbf{Medical Segmentation Decathlon - Prostate} \cite{antonelli2022medical}, composed by 32  prostate MRIs for training, with respective masks of peripheral and transition zone, merged into a single mask for the whole prostate. This dataset was hosted at \emph{Institut Curie} (CURIE).
    \item \textbf{Promise12} \cite{litjens2014evaluation}, consisting of 50 training cases obtained with different scanners. Of those, 27 cases were acquired without using an endorectal coil. This dataset was hosted at \emph{Centre Henri Becquerel} (CHB).
    \item \textbf{ProstateX} \cite{armato2018prostatex}, containing prostate MRIs acquired by using two different scanners, providing prostate segmentation masks for 189 cases \cite{cuocolo2021quality}. This dataset was hosted at \emph{Centre Antoine Lacassagne} (CAL).
\end{itemize}

Each of the three hospitals was assigned a single dataset from the ones listed above, which was subsequently loaded into the Fed-BioMed node client. 
At each site, the data was further divided in a training subset (90\% of samples) and a holdout subset (10\% of samples).
The resulting federated learning setup reflects a realistic scenario presenting the typical data heterogeneity of multi-centric medical imaging applications (see Figure~\ref{fig:experiment-histogram-pixel-intensity} and Appendix~\ref{sec:usecase-suppl} for more details). 
The use of publicly available data is in the scope of this experiment, and is aimed at demonstrating the FL task on a reproducible benchmark, allowing the comparison of the results with respect to the centralized scenario. 
Moreover, this setup facilitated the ethical approval of this experiment by each clinical center, avoiding the cumbersome process related to the use of hospital data. 

\subsubsection{Experimental details}

Within this setup, training hyperparameters were previously identified via a simulated FL scenario on the same data. 
The chosen aggregation method used was FedAvg~\cite{mcmahan2016communication}, and the \texttt{TrainingPlanApproval} security feature was enabled. 
On the hospitals node side, the Fed-BioMed nodes were running on GPU-enabled machines, even though the existence of specific hardware is not a requirement for running Fed-BioMed. 
The central aggregator and the network component were hosted on a separate server provided by Inria. 
The training experiment consisted of 40 rounds with 25 local gradient updates each.
Additional details are provided in the Supplementary Material~\ref{sec:usecase-suppl}.

\subsubsection{FL does not affect final model's performance}
After training, the model obtained a Dice score of $0.868$.
To assess the generalizability of this result, we performed a 5-fold cross-validation with similar 90-10 splits in both a centralized and a simulated federated setting.
The cross-validation Dice score for the simulated FL model was $0.854 \pm 0.028$, while for the centralized (CL) model it was $0.850 \pm 0.035$.
The difference in cross-validation scores between the simulated FL and CL models is not statistically significant ($p=0.63$).
Furthermore, the original model trained in the real-world setting has a realistic performance that falls within $0.5\sigma$ of the average of the simulated FL cross-validation scores.
The distribution of Dice scores for all cross-validation folds combined, shown in Figure~\ref{fig:dsc-score-comparison}, are qualitatively similar between the CL and FL models.  
The final segmentations are also visually indistinguishable, as shown in the bottom row of Figure~\ref{fig:example-seg}.

\subsubsection{FL runtime overhead}
\begin{figure}
    \centering
    \subfloat[]{
        \includegraphics[width=0.495\textwidth]{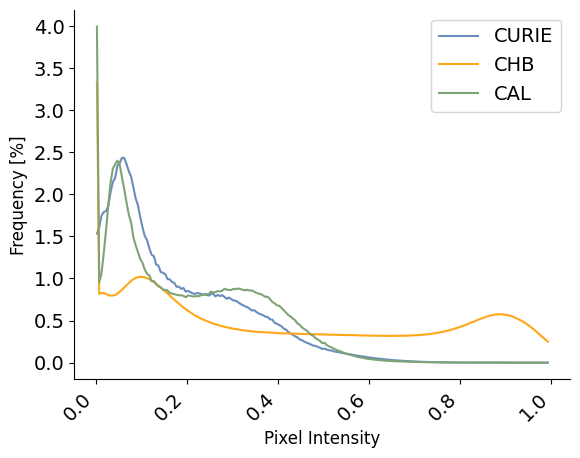}\label{fig:experiment-histogram-pixel-intensity}
    }
    \subfloat[]{
        \includegraphics[width=0.495\textwidth]{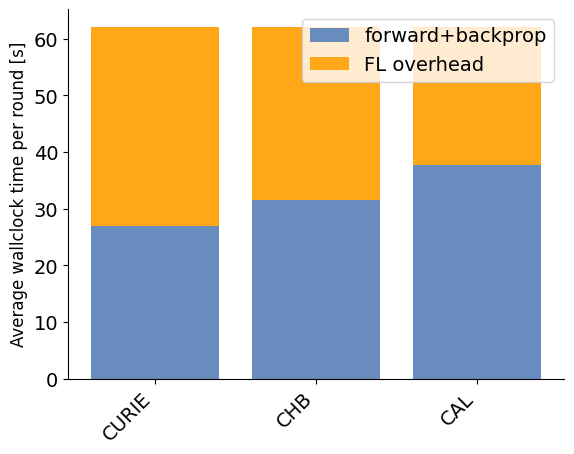}\label{fig:experiment-report-overhead}
    } \\
    \subfloat[]{
        \includegraphics[width=0.495\textwidth]{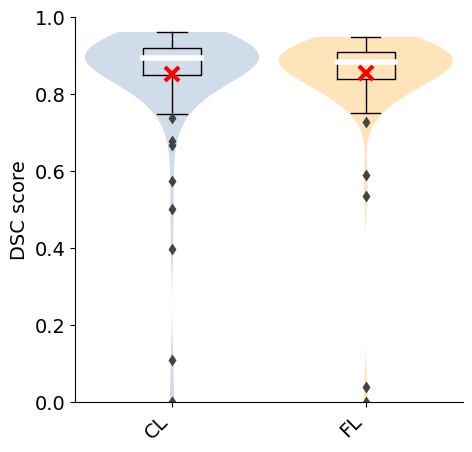}\label{fig:dsc-score-comparison}
    }
    \subfloat[]{
        \includegraphics[width=0.495\textwidth]{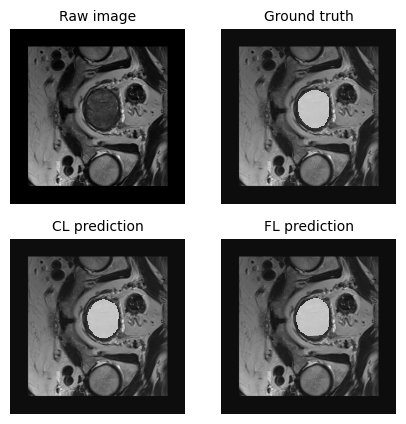}\label{fig:example-seg}
    }
    \caption{\textbf{a)} Pixel intensity distribution grouped by clinical site. 
Prostate images exhibit significant differences between sites, especially in the case of Site 2. 
\textbf{b)} Breakdown of FL experiment wallclock runtime. 
FL introduces a significant overhead in our case, likely attributable to the relatively small number of samples seen by the model in each FL round. 
\textbf{c)} Distribution of Dice scores for centralized (CL) and federated (FL) models, combining all cross-validation folds. 
The performance of the two models has a statistically significant difference, but with a small effect size. 
The figure shows boxplots inside the violin plots. 
The top bottom of each boxplot depict the 3rd and 1st quartile of each measure. 
The white line and the red x indicate the median and mean values, respectively. 
The whiskers depict the extremal observations still within 1.5 times the interquartile range.
\textbf{d)} Clockwise from top left: an example raw image, the ground truth segmentation, the FL model prediction and the CL model prediction.}
\end{figure}

Each round required on average just over 60 seconds of wallclock time for completion.
Federated Learning was found to induce a significant overhead on training time, ranging from 39\% to 56\% of the overall experiment wallclock time, as shown in Figure~\ref{fig:experiment-report-overhead}.
This is in contrast with other studies that report negligible overhead~\cite{truhn2022encrypted,saldanha2022swarm}, and can be explained by the smaller dataset sizes in our experiment, associated with a smaller number of samples used for training within each round. 
We believe that our experiment represents a realistic common scenario for most real-world hospital deployments where data may be scarce and high performance computing not applicable.
Some \fbm implementation details, such as an hard-coded delay at round initialization, may also impact this measure and we are investigating whether useful runtime gains could be made through software improvements.

\section{Lessons learned}
%

\paragraph{Answering to the specific needs of biomedical research applications}
The field of biomedical research is characterized by specific needs that do not necessarily generalize to other fields of application of FL, such as tighter regulations for data protection, very large degrees of heterogeneity, and others identified in Section~\ref{sec:requirements}.
FL frameworks, on the other hand, are often developed with a generic approach in mind, striving for generality in terms of application domains. 
While this approach ensures that such frameworks can be more widely used, this may lead to a sub-optimal experience for users who are interested only in a specific application domain. 
We believe that some communities, and in particular the biomedical research field, may greatly benefit from personalized framework development approaches able to cater to specific requirements such as providing the necessary governance tools, ensuring their usability by members of the application domain's community, and providing a software architecture tailored to the privacy or performance needs. 

\paragraph{Model development and debugging in a federated setting}
The development of a new ML model almost always requires an initial exploratory phase where data are examined, data preprocessing pipelines are designed, the model architecture is refined, and finally hyperparameters are tuned.
Most of the strategies for such operations usually involve some data manipulation and observation, for example it may be insightful to examine the data samples with the highest losses for a given model. 
In a federated setting, all of this is not possible because data are not allowed to travel outside of the source institution. 
This leads to a mismatch between who has the right to examine the data --- i.e. the clinical data sources ---, and who has the knowledge to interpret it --- i.e. the data scientists ---. 
\fbm is designed to operate in high-trust environments where the communication between these two entities is encouraged, therefore we may imagine a scenario where the data scientist instructs an operator on the clinical side to look at specific images and try to identify abnormal patterns. 
Furthermore, our modular \texttt{TrainingPlan} design can support advanced explainable techniques, while still guaranteeing node's data privacy through our \texttt{TrainingPlan} approval process. 
Despite these mitigation efforts, the issue of remotely debugging ML models in a privacy-preserving way remains a difficult problem to solve.

\paragraph{Data preparation prior to federated training}
Data in clinical databases are typically stored in proprietary formats that are not generally readily usable for ML analysis. 
However, all FL frameworks, including \fbm, make the assumption that the input data is provided in a format ready for ingestion by an ML model, or maybe a preprocessing pipeline. 
This gap has been, in our experience, a major source of frustration and delays, with the burden of data export and conversion falling usually on clinical data managers lacking the budget and training for such operations. 
\fbm tries to mitigate this by providing \texttt{Dataset} classes that can be populated with data that has been exported but not highly transformed, for example our \texttt{MedicalFolderDataset} class that can handle data in a simplified BIDS structure. 
In the future, we also plan to extend list of supported formats with some that are closer to the raw data exported by hospital IT systems, such as DICOM for imaging, OMOP or FHIR for clinical data, VCF files for genomics, and others. 
An ideal scenario would include a direct integration with the hospital PACS or EHR systems, however the proprietary nature of most of this software and the strict security rules unfortunately make this an unlikely scenario. 
Regardless of the feasibility, we believe that FL frameworks targeted at biomedical research applications must simplify the process of data preparation and be able to interoperate with standard data formats in their raw form, to ease the burden on clinical data managers and improve data reusability. 

\paragraph{Data privacy and regulation for real-world deployment}
Contrary to controlled academic environments, real-world deployment scenarios are characterized by strict privacy and security measures. 
Some notable examples include: avoiding training if a client's dataset has too few samples, requiring secure aggregation and differential privacy measures, restricting the data flow in the network, and ensuring compliance with hospitals firewall policies.
The integration in \fbm of such security measures to meet the requirements asked by Data Protection Officers of hospitals has been a long, iterative process consisting of multiple rounds of discussion needed to achieve a common understanding and a shared vocabulary.
Ultimately, this dialectic process is what allowed us to carry out our first deployment in a collaborative clinical project.

\section{Conclusion}
We have presented version v4.1 of \fbm, a Python-based FL framework aimed at supporting real-world deployments for biomedical and healthcare research applications.
This document focuses on \fbm's philosophy, guiding principles, and relevant details of the current architecture.
\fbm is an open-source project available in our public repository\footnote{\url{https://gitlab.inria.fr/fedbiomed/fedbiomed}} under an Apache license, and we welcome contributions from the community. 
We also provide extensive user documentation and tutorials, in addition to an API reference for developers\footnote{\url{https://fedbiomed.gitlabpages.inria.fr/v4.1/getting-started/what-is-fedbiomed/}}.
The development of \fbm is an ongoing effort currently brought forward by a group of vetted collaborators coordinated by a small number of core developers. 
In the future, we plan to grow our community of users and developers, and integrate open-source contributions through a consortium.
Our short-term implementation roadmap includes finalizing our implementation of Secure Aggregation, securing client-server communications, and improving even further the usability for clinical data providers.
The domain of healthcare and clinical research is characterized by specific challenges and requirements that are difficult or impossible to satisfy with the currently available frameworks. 
We hope that \fbm will lower the threshold to apply FL in this domain, enabling clinical data scientists to perform experiments and train ML models within a federated consortium of data providers in a secure, practical, and privacy-preserving way.

\FloatBarrier
\section{Supplementary Information}

\subsection{Real-world use case scenario: prostate segmentation}\label{sec:usecase-suppl}

Data were gathered following the process described in~\cite{innocenti2023consensus} from three public datasets~\cite{antonelli2022medical,litjens2014evaluation,armato2018prostatex}. 
Each clinical site received data corresponding to a single source, to reflect the more realistic scenario of high heterogeneity of data among clinical sites.
Details of the data distribution are given in Table~\ref{tab:data-distribution}, while details of the experiment setup are given in Table~\ref{tab:exp-setup}.
We used \fbm v4.1.2 relied on MONAI's (v1.0.1) implementation of UNet~\cite{kerfoot2019left} and data transformations, and on Pytorch's SGD implementation, for definition of the model and optimizer in each site's local training.

\begin{table}[]
    \centering
    \begin{tabular}{c|lcc}
        Center & Dataset & Train samples & holdout samples  \\ \hline 
        CAL & ProstateX~\cite{armato2018prostatex} & 147 & 37 \\
        CHB & Promise~\cite{litjens2014evaluation} & 21 & 6 \\
        CURIE & Decathlon~\cite{antonelli2022medical} & 25 & 7 \\
    \end{tabular}
    \caption{Distribution of data among clinical sites during the FL experiment, including the number of samples in each cross-validation fold.}
    \label{tab:data-distribution}
\end{table}

\begin{table}[]
    \centering
    \begin{tabular}{ll|l}
        \multirow{3}{*}{Data}  & Transformation & Center cropping and padding \\
              & Common shape & 320, 320, 16 \\
              & Transformation & Intensity normalization \\ \hline
        \multirow{5}{*}{UNet} & channels & 16, 32, 64, 128, 256 \\
              & strides & 2, 2, 2, 2 \\
              & residual units & 3 \\
              & normalization & Batch normalization \\
              & dropout & 0.3 \\ \hline
        \multirow{6}{*}{Local} & Optimizer & SGD \\
              & Learning rate & 0.1 \\
              & Momentum & 0.9 \\
              & Batch size & 8 \\
              & Local updates & 25 \\ 
              & Loss & Dice loss \\ \hline
        \multirow{2}{*}{Federated} & Aggregation & FedAVG \\
                  & Rounds & 40 \\
    \end{tabular}
    \caption{FL experimental setup details.}
    \label{tab:exp-setup}
\end{table}


\subsection{Technical implementation details}\label{si:technical}

Building on our functional description in Section~\ref{sec:design-and-functional}, we provide here a more detailed explanation of \fbm's implementation details, as shown in Figure~\ref{fig:fbm-pillars}, trying to make a clear distinction between those that, as of today, have already reached a satisfactory level of maturity and those that are expected to improve in the near future, or may change at some point due to the dynamic nature of our project. 

\subsubsection{Core library functionalities}
On the node side we have implemented: a database of dataset metadata based on the TinyDB\footnote{\url{https://pypi.org/project/tinydb/}} package; a task manager based on persist-queue\footnote{\url{https://github.com/peter-wangxu/persist-queue}}; a suite of \texttt{Dataset} classes to represent tabular, medical imaging (BIDS~\cite{gorgolewski2016brain}), and other data formats; and a \texttt{Round} class handling all the logic for the execution of one federated training round. 
In the future, we plan to vastly extend our library of supported data formats to include other imaging formats such as e.g. DICOM, add support for NLP formats, and integrate interoperability standards and data models such as i2b2 and OMOP\cite{mustra2008overview,murphy2010serving,observational2019book}.
Furthermore, we may consider moving to more complex database solutions if dataset management becomes too cumbersome, and similarly we may consider more advanced task management systems should the need arise among clinical data providers for better resource handling and improved performance.  
On the researcher side, an \texttt{Experiment} class represents the entry-point for the configuration and steering of a FL experiment, while the \texttt{TrainingPlan} class contains all the logic and code to be shipped to nodes for remote execution. 
The highly interactive design of the researcher-facing classes represents one of the highlights of \fbm, and in the future we plan to improve this feature by moving towards an ever more modular design with clearer separation of responsibilities, a hierarchical approach, and improved user documentation. 

Communication between the researcher and the nodes happens through an intermediate component called the \texttt{Network}. 
This component is based on two technologies: MQTT for the brokering of short messages (e.g. train, search datasets, and others)~\cite{banks2019mqtt}, and an HTTP API based on Django REST\footnote{\url{https://www.django-rest-framework.org/}} for exchanging larger files such as model parameters. 
MQTT was chosen because of its natural mapping to a star topology and good support of broadcast operations, which we use for discovering datasets and clients. 
However, in the future we plan to evaluate other approaches, more commonly used in the literature, such as relying on gRPC or other remote execution protocols.
In the current implementation, nodes are pure slaves that execute commands issued by the researcher. 
In the future, we plan to evaluate endowing nodes with a more active role, for example inverting the direction of communication such that the nodes would request tasks instead of passively receiving them, which would entail security benefits (reduced attack surface on the nodes), while shifting some of the complexity from the node implementation to the network component. 

\subsubsection{User interfaces}
In terms of user interfaces, on the node side we have implemented both a Command Line Interface (CLI) and a Graphical User Interface (GUI).
Our GUI implementation is composed of a web client based on React, and a Flask backend server that manages the interactions with the \fbm library, in particular with the dataset database.
This simplifies the process of managing datasets (CRUD operations~\cite{martin1983managing}), training plan approval by data science experts on the node side, and has the general goal of facilitating governance operations on data and models. 
Currently, the GUI is limited to executing on the same machine as the node process, but in the future we plan to decouple the two through a micro-services approach, as well as improve the RBAC with better security and account management.
Another feature that we plan to implement in the future is adding experiment monitoring functionalities for the node, as well as allowing basic experiment steering functionalities such as stopping an experiment through the GUI.
The researcher's interface is a Python SDK, mainly designed to be executed by interpreter such as Jython and within notebooks. 
The researcher may monitor the training via Tensorboard, which plots common metrics such as training loss as well as custom ones defined via a plugin system. 
Defining the best interface for the researcher in terms of tradeoff between interactivity, security, and simplicity is one of our main goals, and we constantly strive to improve the API for the \texttt{Experiment} and \texttt{TrainingPlan} classes. 

\subsubsection{Cybersecurity and model security}
Cybersecurity is one of the main design pillars of \fbm. 
As described above, we have implemented a VPN deployment mode based on the Wireguard framework to isolate the execution of \fbm from external parties~\cite{donenfeld2017wireguard}.
The VPN deployment mode is shipped with the software, and designed to be easily configurable and deployable.
One of our short-term goals is to implement TLS-encrypted communication within the VPN, by exploring suitable MQTT and Django-REST extensions. 
To protect nodes from the execution of arbitrary code, a training plan approval mode can be enabled. 
This process is handled by the GUI on the Node side, making it easy and immediate for clinical data providers to conduct their reviews. 
In this configuration, when a researcher sends a train request to the node, the hash of the training plan is compared to the hashes of a set of training plans that have been previously reviewed and approved by the node, and is executed only in the case of a match.
Differential Privacy can also be enabled in \fbm, when using PyTorch as the training backend, by leveraging the Opacus library and Tensorflow's accountant\footnote{\url{https://raw.githubusercontent.com/tensorflow/privacy/7eea74a6a1cf15e2d2bd890722400edd0e470db8/research/hyperparameters_2022/rdp_accountant.py}}.
Secure aggregation is being implemented based on additively homomorphic encryption~\cite{joye2013scalable,mansouri2022learning}, with the computation of keys based on a multi-party computation (MPC) approach inspired by the MP-SPDZ benchmark suite~\cite{keller2020mpspdz}.

\printbibliography

@article{cuocolo2021quality,
  title={Quality control and whole-gland, zonal and lesion annotations for the PROSTATEx challenge public dataset},
  author={Cuocolo, Renato and Stanzione, Arnaldo and Castaldo, Anna and De Lucia, Davide Raffaele and Imbriaco, Massimo},
  journal={European Journal of Radiology},
  volume={138},
  pages={109647},
  year={2021},
  publisher={Elsevier}
}

@article{dayan2021federated,
  title={Federated learning for predicting clinical outcomes in patients with COVID-19},
  author={Dayan, Ittai and Roth, Holger R and Zhong, Aoxiao and Harouni, Ahmed and Gentili, Amilcare and Abidin, Anas Z and Liu, Andrew and Costa, Anthony Beardsworth and Wood, Bradford J and Tsai, Chien-Sung and others},
  journal={Nature medicine},
  volume={27},
  number={10},
  pages={1735--1743},
  year={2021},
  publisher={Nature Publishing Group}
}

@article{rieke2020future,
  title={The future of digital health with federated learning},
  author={Rieke, Nicola and Hancox, Jonny and Li, Wenqi and Milletari, Fausto and Roth, Holger R and Albarqouni, Shadi and Bakas, Spyridon and Galtier, Mathieu N and Landman, Bennett A and Maier-Hein, Klaus and others},
  journal={NPJ digital medicine},
  volume={3},
  number={1},
  pages={1--7},
  year={2020},
  publisher={Nature Publishing Group}
}

@article{topol2019high,
  title={High-performance medicine: the convergence of human and artificial intelligence},
  author={Topol, Eric J},
  journal={Nature medicine},
  volume={25},
  number={1},
  pages={44--56},
  year={2019},
  publisher={Nature Publishing Group}
}

@article{cuggia2019french,
  title={The French Health Data Hub and the German Medical Informatics Initiatives: two national projects to promote data sharing in healthcare},
  author={Cuggia, Marc and Combes, St{\'e}phanie},
  journal={Yearbook of medical informatics},
  volume={28},
  number={01},
  pages={195--202},
  year={2019},
  publisher={Georg Thieme Verlag KG}
}

@article{sudlow2015uk,
  title={UK biobank: an open access resource for identifying the causes of a wide range of complex diseases of middle and old age},
  author={Sudlow, Cathie and Gallacher, John and Allen, Naomi and Beral, Valerie and Burton, Paul and Danesh, John and Downey, Paul and Elliott, Paul and Green, Jane and Landray, Martin and others},
  journal={PLoS medicine},
  volume={12},
  number={3},
  pages={e1001779},
  year={2015},
  publisher={Public Library of Science}
}

@article{jack2008alzheimer,
  title={The Alzheimer's disease neuroimaging initiative (ADNI): MRI methods},
  author={Jack Jr, Clifford R and Bernstein, Matt A and Fox, Nick C and Thompson, Paul and Alexander, Gene and Harvey, Danielle and Borowski, Bret and Britson, Paula J and L. Whitwell, Jennifer and Ward, Chadwick and others},
  journal={Journal of Magnetic Resonance Imaging: An Official Journal of the International Society for Magnetic Resonance in Medicine},
  volume={27},
  number={4},
  pages={685--691},
  year={2008},
  publisher={Wiley Online Library}
}

@article{tomczak2015cancer,
  title={The cancer genome atlas (tcga): an immeasurable source of knowledge.},
  author={Tomczak, K. and Czerwińska, P. and Wiznerowicz, M.},
  journal={Contemporary Oncology},
  volume={19},
  year={2015}
}

@article{sporns2005human,
  title={The human connectome: a structural description of the human brain},
  author={Sporns, Olaf and Tononi, Giulio and K{\"o}tter, Rolf},
  journal={PLoS computational biology},
  volume={1},
  number={4},
  pages={e42},
  year={2005},
  publisher={Public Library of Science San Francisco, USA}
}

@article{menze2014multimodal,
  title={The multimodal brain tumor image segmentation benchmark (BRATS)},
  author={Menze, Bjoern H and Jakab, Andras and Bauer, Stefan and Kalpathy-Cramer, Jayashree and Farahani, Keyvan and Kirby, Justin and Burren, Yuliya and Porz, Nicole and Slotboom, Johannes and Wiest, Roland and others},
  journal={IEEE transactions on medical imaging},
  volume={34},
  number={10},
  pages={1993--2024},
  year={2014},
  publisher={IEEE}
}

@article{clark2013cancer,
  title={The Cancer Imaging Archive (TCIA): maintaining and operating a public information repository},
  author={Clark, Kenneth and Vendt, Bruce and Smith, Kirk and Freymann, John and Kirby, Justin and Koppel, Paul and Moore, Stephen and Phillips, Stanley and Maffitt, David and Pringle, Michael and others},
  journal={Journal of digital imaging},
  volume={26},
  number={6},
  pages={1045--1057},
  year={2013},
  publisher={Springer}
}

@article{van2014systematic,
  title={A systematic review of barriers to data sharing in public health},
  author={Van Panhuis, Willem G and Paul, Proma and Emerson, Claudia and Grefenstette, John and Wilder, Richard and Herbst, Abraham J and Heymann, David and Burke, Donald S},
  journal={BMC public health},
  volume={14},
  number={1},
  pages={1--9},
  year={2014},
  publisher={BioMed Central}
}

@article{mcmahan2016communication,
  title={Communication-efficient learning of deep networks from decentralized data. arXiv e-prints},
  author={McMahan, H and Moore, Eider and Ramage, Daniel and Hampson, Seth and Aguera y Arcas, B},
  journal={Artificial intelligence and statistics. PMLR.},
  year={2017}
}

@article{crowson2022systematic,
  title={A systematic review of federated learning applications for biomedical data},
  author={Crowson, Matthew G and Moukheiber, Dana and Ar{\'e}valo, Aldo Robles and Lam, Barbara D and Mantena, Sreekar and Rana, Aakanksha and Goss, Deborah and Bates, David W and Celi, Leo Anthony},
  journal={PLOS Digital Health},
  volume={1},
  number={5},
  pages={e0000033},
  year={2022},
  publisher={Public Library of Science San Francisco, CA USA}
}

@article{darzidehkalani2022federated1,
  title={Federated Learning in Medical Imaging: Part I: Toward Multicentral Health Care Ecosystems},
  author={Darzidehkalani, Erfan and Ghasemi-Rad, Mohammad and van Ooijen, Pma},
  journal={Journal of the American College of Radiology},
  year={2022},
  publisher={Elsevier}
}

@article{shaheen2022applications,
  title={Applications of federated learning; Taxonomy, challenges, and research trends},
  author={Shaheen, Momina and Farooq, Muhammad Shoaib and Umer, Tariq and Kim, Byung-Seo},
  journal={Electronics},
  volume={11},
  number={4},
  pages={670},
  year={2022},
  publisher={MDPI}
}

@article{sadilek2021privacy,
  title={Privacy-first health research with federated learning},
  author={Sadilek, Adam and Liu, Luyang and Nguyen, Dung and Kamruzzaman, Methun and Serghiou, Stylianos and Rader, Benjamin and Ingerman, Alex and Mellem, Stefan and Kairouz, Peter and Nsoesie, Elaine O and others},
  journal={NPJ digital medicine},
  volume={4},
  number={1},
  pages={1--8},
  year={2021},
  publisher={Nature Publishing Group}
}

@article{hulsen2019big,
  title={From big data to precision medicine},
  author={Hulsen, Tim and Jamuar, Saumya S and Moody, Alan R and Karnes, Jason H and Varga, Orsolya and Hedensted, Stine and Spreafico, Roberto and Hafler, David A and McKinney, Eoin F},
  journal={Frontiers in medicine},
  pages={34},
  year={2019},
  publisher={Frontiers}
}

@article{murphy2010serving,
  title={Serving the enterprise and beyond with informatics for integrating biology and the bedside (i2b2)},
  author={Murphy, Shawn N and Weber, Griffin and Mendis, Michael and Gainer, Vivian and Chueh, Henry C and Churchill, Susanne and Kohane, Isaac},
  journal={Journal of the American Medical Informatics Association},
  volume={17},
  number={2},
  pages={124--130},
  year={2010},
  publisher={BMJ Group}
}

@article{gorgolewski2016brain,
  title={The brain imaging data structure, a format for organizing and describing outputs of neuroimaging experiments},
  author={Gorgolewski, Krzysztof J and Auer, Tibor and Calhoun, Vince D and Craddock, R Cameron and Das, Samir and Duff, Eugene P and Flandin, Guillaume and Ghosh, Satrajit S and Glatard, Tristan and Halchenko, Yaroslav O and others},
  journal={Scientific data},
  volume={3},
  number={1},
  pages={1--9},
  year={2016},
  publisher={Nature Publishing Group}
}

@inproceedings{mustra2008overview,
  title={Overview of the DICOM standard},
  author={Mustra, Mario and Delac, Kresimir and Grgic, Mislav},
  booktitle={2008 50th International Symposium ELMAR},
  volume={1},
  pages={39--44},
  year={2008},
  organization={IEEE}
}

@misc{european2019ict,
  title={ICT Security Certification Opportunities in the Healthcare Sector},
  author={European Union Agency for Cybersecurity (ENISA)},
  year={2018},
  publisher={European Union Agency For Network and Information Security Attiki, Greece},
  volume={1.0}
}

@article{galtier2019substra,
  title={Substra: a framework for privacy-preserving, traceable and collaborative machine learning},
  author={Galtier, Mathieu N and Marini, Camille},
  journal={arXiv preprint arXiv:1910.11567},
  year={2019}
}

@article{pati2022federated,
  title={Federated Learning Enables Big Data for Rare Cancer Boundary Detection},
  author={Pati, Sarthak and Baid, Ujjwal and Edwards, Brandon and Sheller, Micah and Wang, Shih-Han and Reina, G Anthony and Foley, Patrick and Gruzdev, Alexey and Karkada, Deepthi and Davatzikos, Christos and others},
  journal={Nature Communication},
  year={2022}
}

@article{ogier2023federated,
  title={Federated learning for predicting histological response to neoadjuvant chemotherapy in triple-negative breast cancer},
  author={Ogier du Terrail, Jean and Leopold, Armand and Joly, Cl{\'e}ment and B{\'e}guier, Constance and Andreux, Mathieu and Maussion, Charles and Schmauch, Beno{\^\i}t and Tramel, Eric W and Bendjebbar, Etienne and Zaslavskiy, Mikhail and others},
  journal={Nature Medicine},
  pages={1--12},
  year={2023},
  publisher={Nature Publishing Group US New York}
}

@article{beutel2020flower,
  title={Flower: A friendly federated learning research framework},
  author={Beutel, Daniel J and Topal, Taner and Mathur, Akhil and Qiu, Xinchi and Parcollet, Titouan and de Gusm{\~a}o, Pedro PB and Lane, Nicholas D},
  journal={arXiv preprint arXiv:2007.14390},
  year={2020}
}

@article{he2020fedml,
  title={Fedml: A research library and benchmark for federated machine learning},
  author={He, Chaoyang and Li, Songze and So, Jinhyun and Zeng, Xiao and Zhang, Mi and Wang, Hongyi and Wang, Xiaoyang and Vepakomma, Praneeth and Singh, Abhishek and Qiu, Hang and others},
  journal={arXiv preprint arXiv:2007.13518},
  year={2020}
}

@article{ludwig2020ibm,
  title={Ibm federated learning: an enterprise framework white paper v0. 1},
  author={Ludwig, Heiko and Baracaldo, Nathalie and Thomas, Gegi and Zhou, Yi and Anwar, Ali and Rajamoni, Shashank and Ong, Yuya and Radhakrishnan, Jayaram and Verma, Ashish and Sinn, Mathieu and others},
  journal={arXiv preprint arXiv:2007.10987},
  year={2020}
}

@article{bonawitz2019towards,
  title={Towards federated learning at scale: System design},
  author={Bonawitz, Keith and Eichner, Hubert and Grieskamp, Wolfgang and Huba, Dzmitry and Ingerman, Alex and Ivanov, Vladimir and Kiddon, Chloe and Kone{\v{c}}n{\`y}, Jakub and Mazzocchi, Stefano and McMahan, Brendan and others},
  journal={Proceedings of Machine Learning and Systems},
  volume={1},
  pages={374--388},
  year={2019}
}

@techreport{owkin2022substrafl,
  title={Substrafl overview},
  author={Owkin},
  year={2022},
  url={https://docs.substra.org/en/stable/substrafl_doc/substrafl_overview.html#},
  note={Accessed on: 28th November 2022, Revision b152eecf}
}

@techreport{owkin2019mellody,
  title={Mellody},
  author={Mellody consortium},
  year={2019},
  url={https://www.melloddy.eu/},
  note={Accessed on 28th November 2022}
}

@techreport{owkin2019healthchain,
  title={Healthchain},
  author={Owkin},
  year={2019},
  url={https://www.labelia.org/fr/healthchain-project},
  note={Accessed on 28th November 2022}
}

@article{foley2022openfl,
	author={Foley, Patrick and Sheller, Micah J and Edwards, Brandon and Pati, Sarthak and Riviera, Walter and Sharma, Mansi and Moorthy, Prakash Narayana and Wang, Shi-han and Martin, Jason and Mirhaji, Parsa and Shah, Prashant and Bakas, Spyridon},
	title={OpenFL: the open federated learning library},
	journal={Physics in Medicine \& Biology},
	url={http://iopscience.iop.org/article/10.1088/1361-6560/ac97d9},
	year={2022},
	doi={10.1088/1361-6560/ac97d9},
	publisher={IOP Publishing}
}

@inproceedings{roth2022flare,
  title={FLARE: Federated Learning from Simulation to Real-World},
  author={Roth, Holger R and Cheng, Yan and Wen, Yuhong and Yang, Isaac and Xu, Ziyue and Hsieh, YuanTing and Kersten, Kristopher and Harouni, Ahmed and Zhao, Can and Lu, Kevin and others},
  booktitle={Workshop on Federated Learning: Recent Advances and New Challenges (in Conjunction with NeurIPS 2022)}
}

@incollection{roth2020federated,
  title={Federated learning for breast density classification: A real-world implementation},
  author={Roth, Holger R and Chang, Ken and Singh, Praveer and Neumark, Nir and Li, Wenqi and Gupta, Vikash and Gupta, Sharut and Qu, Liangqiong and Ihsani, Alvin and Bizzo, Bernardo C and others},
  booktitle={Domain adaptation and representation transfer, and distributed and collaborative learning},
  pages={181--191},
  year={2020},
  publisher={Springer}
}

@article{sarma2021federated,
  title={Federated learning improves site performance in multicenter deep learning without data sharing},
  author={Sarma, Karthik V and Harmon, Stephanie and Sanford, Thomas and Roth, Holger R and Xu, Ziyue and Tetreault, Jesse and Xu, Daguang and Flores, Mona G and Raman, Alex G and Kulkarni, Rushikesh and others},
  journal={Journal of the American Medical Informatics Association},
  volume={28},
  number={6},
  pages={1259--1264},
  year={2021},
  publisher={Oxford University Press}
}

@article{deist2020distributed, 
    title={Distributed learning on 20 000+ lung cancer patients – The Personal Health Train}, 
    volume={144}, 
    url={https://linkinghub.elsevier.com/retrieve/pii/S0167814019334899}, 
    DOI={10.1016/j.radonc.2019.11.019}, 
    journal={Radiotherapy and Oncology}, 
    author={Deist, Timo M. and Dankers, Frank J.W.M. and Ojha, Priyanka and Scott Marshall, M. and Janssen, Tomas and Faivre-Finn, Corinne and Masciocchi, Carlotta and Valentini, Vincenzo and Wang, Jiazhou and Chen, Jiayan and Zhang, Zhen and Spezi, Emiliano and Button, Mick and Jan Nuyttens, Joost and Vernhout, René and van Soest, Johan and Jochems, Arthur and Monshouwer, René and Bussink, Johan and Price, Gareth and Lambin, Philippe and Dekker, Andre}, 
    year={2020}, 
    pages={189–200}, 
    language={en} 
}

@article{saldanha2022swarm, 
    title={Swarm learning for decentralized artificial intelligence in cancer histopathology}, 
    volume={28}, 
    url={https://www.nature.com/articles/s41591-022-01768-5}, DOI={10.1038/s41591-022-01768-5}, 
    number={6}, 
    journal={Nature Medicine}, 
    author={Saldanha, Oliver Lester and Quirke, Philip and West, Nicholas P. and James, Jacqueline A. and Loughrey, Maurice B. and Grabsch, Heike I. and Salto-Tellez, Manuel and Alwers, Elizabeth and Cifci, Didem and Ghaffari Laleh, Narmin and Seibel, Tobias and Gray, Richard and Hutchins, Gordon G. A. and Brenner, Hermann and van Treeck, Marko and Yuan, Tanwei and Brinker, Titus J. and Chang-Claude, Jenny and Khader, Firas and Schuppert, Andreas and Luedde, Tom and Trautwein, Christian and Muti, Hannah Sophie and Foersch, Sebastian and Hoffmeister, Michael and Truhn, Daniel and Kather, Jakob Nikolas}, 
    year={2022}, 
    month={6}, 
    pages={1232–1239}, 
    language={en} 
}

@article{froelicher2021truly, 
    title={Truly privacy-preserving federated analytics for precision medicine with multiparty homomorphic encryption}, 
    volume={12}, 
    url={https://www.nature.com/articles/s41467-021-25972-y}, DOI={10.1038/s41467-021-25972-y}, 
    number={1}, 
    journal={Nature Communications}, 
    author={Froelicher, David and Troncoso-Pastoriza, Juan R. and Raisaro, Jean Louis and Cuendet, Michel A. and Sousa, Joao Sa and Cho, Hyunghoon and Berger, Bonnie and Fellay, Jacques and Hubaux, Jean-Pierre}, 
    year={2021}, 
    month={10}, 
    pages={5910}, 
    language={en} 
}

@article{ogundokun2022review, 
  title={A Review on Federated Learning and Machine Learning Approaches: Categorization, Application Areas, and Blockchain Technology}, 
  volume={13}, url={https://www.mdpi.com/2078-2489/13/5/263}, DOI={10.3390/info13050263}, 
  number={5}, 
  journal={Information}, 
  author={Ogundokun, Roseline Oluwaseun and Misra, Sanjay and Maskeliunas, Rytis and Damasevicius, Robertas}, 
  year={2022}, 
  month={5}, 
  pages={263}, 
  language={en} 
}

@article{li2021survey,  
  author={Li, Qinbin and Wen, Zeyi and Wu, Zhaomin and Hu, Sixu and Wang, Naibo and Li, Yuan and Liu, Xu and He, Bingsheng},  
  journal={IEEE Transactions on Knowledge and Data Engineering},   
  title={A Survey on Federated Learning Systems: Vision, Hype and Reality for Data Privacy and Protection},   
  year={2021},  
  volume={},  
  number={},  
  pages={1-1},  
  doi={10.1109/TKDE.2021.3124599}
}

@inproceedings{joye2013scalable,
  title={A scalable scheme for privacy-preserving aggregation of time-series data},
  author={Joye, Marc and Libert, Beno\^{i}t},
  booktitle={International Conference on Financial Cryptography and Data Security},
  pages={111--125},
  year={2013},
  organization={Springer}
}

@conference{mansouri2022learning,
  author = {Mansouri, Mohamad and  Önen, Melek and  Ben Jaballah, Wafa},
  title = {Learning from failures: Secure and fault-tolerant aggregation for federated learning},
  booktitle = {ACSAC 2022, Annual Computer Security Applications Conference, 5-9 December 2022, Austin, Texas, USA},
  year = {2022},
  editor = {ACM},
  address = {Austin},
  note = {© ACM, 2022. This is the author\&\#039;s version of the work. It is posted here by permission of ACM for your personal use. Not for redistribution. The definitive version was published in ACSAC 2022, Annual Computer Security Applications Conference, 5-9 December 2022, Austin, Texas, USA https://doi.org/10.1145/3564625.3568135},
}

@article{ghassemi2020review, 
  title={A Review of Challenges and Opportunities in Machine Learning for Health},
  volume={2020}, 
  url={https://www.ncbi.nlm.nih.gov/pmc/articles/PMC7233077/},
  journal={AMIA Summits on Translational Science Proceedings}, 
  author={Ghassemi, Marzyeh and Naumann, Tristan and Schulam, Peter and Beam, Andrew L. and Chen, Irene Y. and Ranganath, Rajesh}, 
  year={2020}, 
  month={5}, 
  pages={191–200} 
}

@article{kline2022multimodal,
  title={Multimodal machine learning in precision health: A scoping review},
  author={Kline, Adrienne and Wang, Hanyin and Li, Yikuan and Dennis, Saya and Hutch, Meghan and Xu, Zhenxing and Wang, Fei and Cheng, Feixiong and Luo, Yuan},
  journal={npj Digital Medicine},
  volume={5},
  number={1},
  pages={1--14},
  year={2022},
  publisher={Nature Publishing Group}
}

@article{bouacida2021vulnerabilities,
  title={Vulnerabilities in federated learning},
  author={Bouacida, Nader and Mohapatra, Prasant},
  journal={IEEE Access},
  volume={9},
  pages={63229--63249},
  year={2021},
  publisher={IEEE}
}

@article{shyu2021systematic,
  title={A systematic review of federated learning in the healthcare area: From the perspective of data properties and applications},
  author={Shyu, Chi-Ren and Putra, Karisma Trinanda and Chen, Hsing-Chung and Tsai, Yuan-Yu and Hossain, KSM Tozammel and Jiang, Wei and Shae, Zon-Yin},
  journal={Applied Sciences},
  volume={11},
  number={23},
  pages={11191},
  year={2021},
  publisher={MDPI}
}

@article{liu2021fate,
  title={FATE: An Industrial Grade Platform for Collaborative Learning With Data Protection.},
  author={Liu, Yang and Fan, Tao and Chen, Tianjian and Xu, Qian and Yang, Qiang},
  journal={J. Mach. Learn. Res.},
  volume={22},
  number={226},
  pages={1--6},
  year={2021}
}

@article{ma2019paddlepaddle,
  title={PaddlePaddle: An open-source deep learning platform from industrial practice},
  author={Ma, Yanjun and Yu, Dianhai and Wu, Tian and Wang, Haifeng},
  journal={Frontiers of Data and Domputing},
  volume={1},
  number={1},
  pages={105--115},
  year={2019}
}

@article{ryffel2018generic,
  title={A generic framework for privacy preserving deep learning},
  author={Ryffel, Theo and Trask, Andrew and Dahl, Morten and Wagner, Bobby and Mancuso, Jason and Rueckert, Daniel and Passerat-Palmbach, Jonathan},
  journal={arXiv preprint arXiv:1811.04017},
  year={2018}
}

@misc{grpc2018high,
  title={high performance, open-source universal RPC framework},
  author={Foundation, Cloud Native Computing},
  year={2018},
  url={\url{https://grpc.io/}},
  note={Accessed on 16 December 2022}
}

@article{abadi2015tensorflow,
title={ {TensorFlow}: Large-Scale Machine Learning on Heterogeneous Systems},
url={https://www.tensorflow.org/},
note={Software available from tensorflow.org},
author={
    Mart\'{i}n~Abadi and
    Ashish~Agarwal and
    Paul~Barham and
    Eugene~Brevdo and
    Zhifeng~Chen and
    Craig~Citro and
    Greg~S.~Corrado and
    Andy~Davis and
    Jeffrey~Dean and
    Matthieu~Devin and
    Sanjay~Ghemawat and
    Ian~Goodfellow and
    Andrew~Harp and
    Geoffrey~Irving and
    Michael~Isard and
    Yangqing Jia and
    Rafal~Jozefowicz and
    Lukasz~Kaiser and
    Manjunath~Kudlur and
    Josh~Levenberg and
    Dandelion~Man\'{e} and
    Rajat~Monga and
    Sherry~Moore and
    Derek~Murray and
    Chris~Olah and
    Mike~Schuster and
    Jonathon~Shlens and
    Benoit~Steiner and
    Ilya~Sutskever and
    Kunal~Talwar and
    Paul~Tucker and
    Vincent~Vanhoucke and
    Vijay~Vasudevan and
    Fernanda~Vi\'{e}gas and
    Oriol~Vinyals and
    Pete~Warden and
    Martin~Wattenberg and
    Martin~Wicke and
    Yuan~Yu and
    Xiaoqiang~Zheng},
  year={2015},
}

@book{martin1983managing,
  title={Managing the data base environment},
  author={Martin, James},
  year={1983},
  publisher={Prentice Hall PTR}
}

@book{observational2019book,
  title={The Book of OHDSI},
  author={Observational Health Data Sciences and Informatics},
  publisher={Independently published},
  year={2019},
  doi={\url{https://doi.org/10.5281/zenodo.4265255}}
}

@techreport{banks2019mqtt,
  title={MQTT Version 5.0},
  author={Andrew Banks and Ed Briggs and Ken Borgendale and Rahul Gupta},
  year={2019}
}

@inproceedings{donenfeld2017wireguard,
  title={Wireguard: next generation kernel network tunnel},
  author={Donenfeld, Jason A},
  booktitle={NDSS},
  pages={1--12},
  year={2017}
}

@inproceedings{keller2020mpspdz,
    author = {Marcel Keller},
    title = {{MP-SPDZ}: A Versatile Framework for Multi-Party Computation},
    booktitle = {Proceedings of the 2020 ACM SIGSAC Conference on
    Computer and Communications Security},
    year = {2020},
    doi = {10.1145/3372297.3417872},
    url = {https://doi.org/10.1145/3372297.3417872},
}

@article{truhn2022encrypted,
    title={Encrypted federated learning for secure decentralized collaboration in cancer image analysis}, 
    url={https://www.medrxiv.org/content/10.1101/2022.07.28.22277288v1},
    DOI={10.1101/2022.07.28.22277288},
    publisher={medRxiv}, 
    author={Truhn, Daniel and Arasteh, Soroosh Tayebi and Saldanha, Oliver Lester and Müller-Franzes, Gustav and Khader, Firas and Quirke, Philip and West, Nicholas P. and Gray, Richard and Hutchins, Gordon G. A. and James, Jacqueline A. and Loughrey, Maurice B. and Salto-Tellez, Manuel and Brenner, Hermann and Brobeil, Alexander and Yuan, Tanwei and Chang-Claude, Jenny and Hoffmeister, Michael and Foersch, Sebastian and Han, Tianyu and Keil, Sebastian and Schulze-Hagen, Maximilian and Isfort, Peter and Bruners, Philipp and Kaissis, Georgios and Kuhl, Christiane and Nebelung, Sven and Kather, Jakob Nikolas}, 
    year={2022}, 
    month={7}, 
    language={en} 
}

@article{innocenti2023consensus,
    title={Consensus-Based Methods as a Cost-Effective Alternative to Federated Learning: Benchmark on Prostate MRI Segmentation},
    author={Innocenti, Lucia et al.},
    year={2023},
    note={under review}
}

@article{antonelli2022medical,
  title={The medical segmentation decathlon},
  author={Antonelli, Michela and Reinke, Annika and Bakas, Spyridon and Farahani, Keyvan and Kopp-Schneider, Annette and Landman, Bennett A and Litjens, Geert and Menze, Bjoern and Ronneberger, Olaf and Summers, Ronald M and others},
  journal={Nature communications},
  volume={13},
  number={1},
  pages={4128},
  year={2022},
  publisher={Nature Publishing Group UK London}
}

@article{litjens2014evaluation,
  title={Evaluation of prostate segmentation algorithms for MRI: the PROMISE12 challenge},
  author={Litjens, Geert and Toth, Robert and Van De Ven, Wendy and Hoeks, Caroline and Kerkstra, Sjoerd and van Ginneken, Bram and Vincent, Graham and Guillard, Gwenael and Birbeck, Neil and Zhang, Jindang and others},
  journal={Medical image analysis},
  volume={18},
  number={2},
  pages={359--373},
  year={2014},
  publisher={Elsevier}
}

@article{armato2018prostatex,
  title={PROSTATEx Challenges for computerized classification of prostate lesions from multiparametric magnetic resonance images},
  author={Armato III, Samuel G and Huisman, Henkjan and Drukker, Karen and Hadjiiski, Lubomir and Kirby, Justin S and Petrick, Nicholas and Redmond, George and Giger, Maryellen L and Cha, Kenny and Mamonov, Artem and others},
  journal={Journal of Medical Imaging},
  volume={5},
  number={4},
  pages={044501--044501},
  year={2018},
  publisher={Society of Photo-Optical Instrumentation Engineers}
}

@inproceedings{kerfoot2019left,
  title={Left-ventricle quantification using residual U-Net},
  author={Kerfoot, Eric and Clough, James and Oksuz, Ilkay and Lee, Jack and King, Andrew P and Schnabel, Julia A},
  booktitle={Statistical Atlases and Computational Models of the Heart. Atrial Segmentation and LV Quantification Challenges: 9th International Workshop, STACOM 2018, Held in Conjunction with MICCAI 2018, Granada, Spain, September 16, 2018, Revised Selected Papers 9},
  pages={371--380},
  year={2019},
  organization={Springer}
}
\end{document}